\documentclass[journal]{IEEEtran}
\usepackage{amsmath,amsfonts}
\usepackage{algorithmic}
\usepackage{algorithm}
\usepackage{array}
\usepackage[caption=false,font=normalsize,labelfont=sf,textfont=sf]{subfig}
\usepackage{textcomp}
\usepackage{stfloats}
\usepackage{url}
\usepackage{verbatim}
\usepackage{graphicx}
\usepackage{cite}
\usepackage{multirow}
\usepackage{colortbl}
\usepackage{booktabs}
\usepackage{wrapfig}
\usepackage{tcolorbox}
\usepackage{xcolor}
\definecolor{darkpurple}{rgb}{0.5, 0, 0.5}
\definecolor{Highlight}{HTML}{39b54a}
\newcommand{\hl}[1]{\textcolor{Highlight}{#1}}

\usepackage{pifont}
\usepackage{amsmath,graphicx}  
\usepackage[colorlinks=true, linkcolor=blue, citecolor=blue, urlcolor=blue]{hyperref}
\usepackage[nameinlink]{cleveref}  

\begin{document}

\title{PEBench: A Fictitious Dataset to Benchmark Machine Unlearning for Multimodal Large Language Models}




\author{Zhaopan Xu, Pengfei Zhou, Weidong Tang, Jiaxin Ai, Wangbo Zhao, Kai Wang, Wenqi Shao, \\
Xiaojiang Peng,~\IEEEmembership{Senior Member,~IEEE},   Hongxun Yao\dag,~\IEEEmembership{Member,~IEEE}, Kaipeng Zhang\dag,~\IEEEmembership{Member,~IEEE}

\thanks{This work is completed during Zhaopan Xu’s internship at Shanghai Artificial Intelligence Laboratory. This work was supported by the National Key R\&D Program of China No.2022ZD0160102 and the National Science Foundation of China under Grant 62476069. Zhaopan Xu and Hongxun Yao are with the School of Computer, Harbin Institute of Technology, China, 150001. (e-mail: 20b903054@stu.hit.edu.cn; h.yao@hit.edu.cn;).
} 
\thanks{Pengfei Zhou, Wangbo Zhao and Kai Wang are with the School of Computer, National University of Singapore, Singapore. (e-mail: e1374451@u.nus.edu; e0983526@u.nus.edu; kai.wang@comp.nus.edu.sg;).}
\thanks{Weidong Tang is with the School of Computer, Xidian University, Singapore, Xian 710000, China (e-mail: wdtang0705@gmail.com)}
\thanks{Jiaxin Ai is with the School of Computer Science, Wuhan Univeristy, Wuhan 430072, China (e-mail: julyai@whu.edu.cn)}

\thanks{Xiaojiang Peng is with the College of Big Data and Internet, Shenzhen Technology University, Shenzhen, 518118, China. (e-mail: pengxiao jiang@sztu.edu.cn)}
\thanks{Wenqi Shao and Kaipeng Zhang are with Shanghai Artificial Intelligence Laboratory, Shanghai 200000, China (e-mail: shaowenqi@pjlab.orn.cn; zhangkaipeng@pjlab.org.cn).
}
\thanks{\dag Corresponding author.}

}

\maketitle

\begin{abstract}
Multimodal large language models (MLLMs) have achieved remarkable success in vision-language tasks, but their reliance on vast, internet-sourced data raises significant privacy and security concerns. Machine unlearning (MU) has emerged as a critical technique to address these issues, enabling the selective removal of targeted information from pre-trained models without costly retraining. However, the evaluation of MU for MLLMs remains inadequate. Existing benchmarks often lack a comprehensive scope, focusing narrowly on entities while overlooking the unlearning of broader visual concepts and the inherent semantic coupling between them. To bridge this gap, we introduce, PEBench, a novel benchmark designed to facilitate a thorough assessment of MU in MLLMs. PEBench features a fictitious dataset of personal entities and corresponding event scenes to evaluate unlearning across these distinct yet entangled concepts. We leverage this benchmark to evaluate five MU methods, revealing their unique strengths and weaknesses. Our findings show that unlearning one concept can unintentionally degrade performance on related concepts within the same image, a challenge we term cross-concept interference. Furthermore, we demonstrate the difficulty of unlearning person and event concepts simultaneously and propose an effective method to mitigate these conflicting objectives. The source code and benchmark are publicly available at https://pebench.github.io. 
\end{abstract}

\section{Introduction}
\label{sec:intro}

With the rapid development and widespread application of large language models (LLMs), ethical and safety concerns have drawn much attention, due to the large volumes of data scraped from the internet during the training~\cite{bender2021dangers,kotek2023gender}. In response, Machine Unlearning (MU) has been proposed as a remedy~\cite{cao2015towards,zhou2025truvrf,li2025machine}. MU is designed to selectively remove the influence and impact of undesirable data from pre-trained models without requiring complete retraining~\cite{chen2025fedmua}. Recent progress continues to highlight its increasing effectiveness in addressing these challenges within LLMs~\cite{liu2024large,patil2023can}.

\begin{figure}[t]
    \centering 
    \includegraphics[width=0.98\columnwidth]{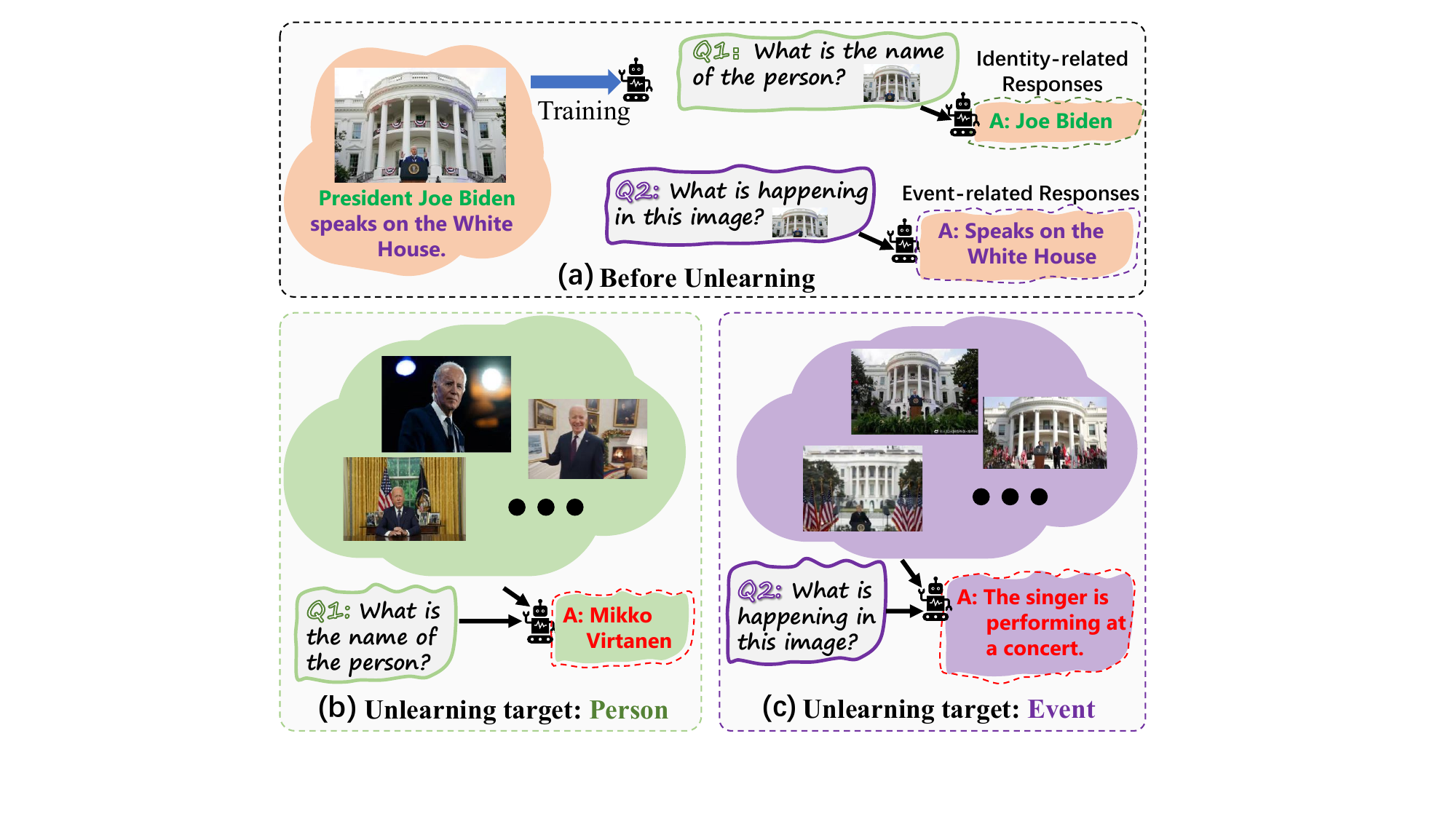}
    \vspace{-5pt}
    \caption{Example of an image of Joe Biden speaking at the White House. Before unlearning (a), MLLMs can generate responses related to various visual concepts (\textcolor{Highlight}{Person} and \textcolor{darkpurple}{Event}). The goal of Machine Unlearning (MU) for MLLMs is to selectively forget specific concepts within the model. When the \textcolor{Highlight}{unlearning target is Person} (b), the model mistakenly \textcolor{red}{identifies Joe Biden as a different person}. When the \textcolor{darkpurple}{unlearning target is Event} (c), the model \textcolor{red}{misinterprets the speech as a concert}.}
    \label{fig:figure1}
    \vspace{-5pt}  
\end{figure}

Building on the foundation of LLMs, Multimodal Large Language Models (MLLMs) have achieved remarkable performance in multimodal applications~\cite{zhang2024mm,huang2024survey,zhang2025modality}. Consequently, MLLMs inherit the same privacy and security vulnerabilities, including risks of copyright infringement and the retention of sensitive information embedded in their training data~\cite{karamolegkou2023copyright,huang2024demystifying}. Developing effective machine unlearning (MU) is therefore equally critical for MLLMs.

\begin{figure*}[t] \centering
\centering
  \resizebox{1\linewidth}{!} {
    \includegraphics{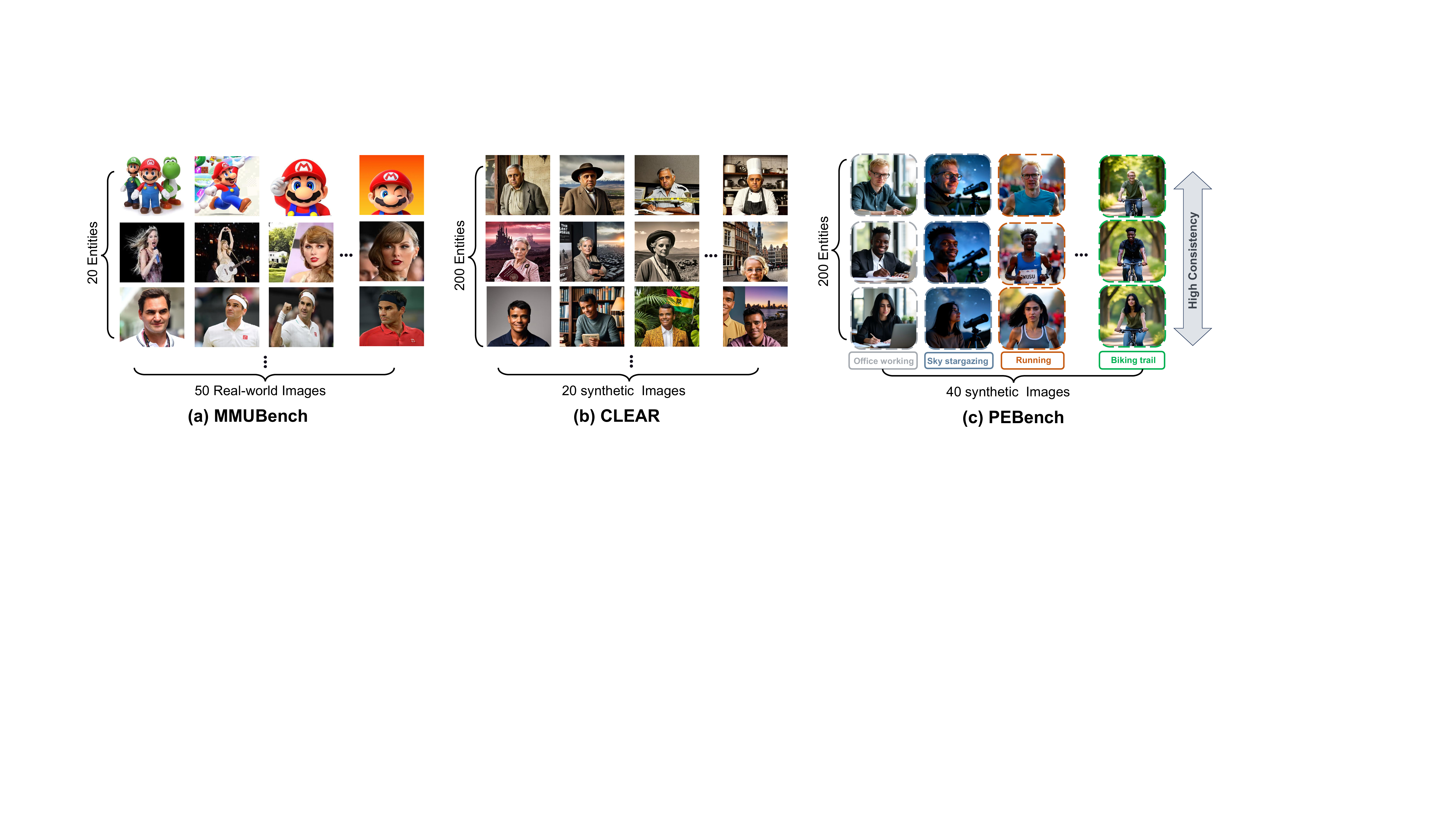}
  }
  \caption{Comparison between previous MU benchmarks and our PEBench for MLLMs. MMUBench (a) utilizes 50 real-world images representing 20 distinct entities, such as popular characters like Mario. CLEAR (b) employs 20 synthetic images for 200 fictitious identities, extending the paradigm of using fictitious data. In contrast, our proposed PEBench (c) features synthetic images encompassing 200 fictitious entities across 40 distinct event scenes, designed to ensure high stylistic consistency across different individuals and various contexts.}
    \label{fig:image2}
  \vspace{-5pt}
\end{figure*}

While numerous benchmarks have been proposed to advance MU in LLMs~\cite{eldan2023s,maini2024tofu}, the landscape for MLLMs remains underdeveloped. Existing MLLM-specific benchmarks, such as MMUBench~\cite{li2024single} (Fig. \ref{fig:image2}(a)) and CLEAR~\cite{dontsov2024clear}  (Fig. \ref{fig:image2}(b)), are constrained to unlearning entities. This narrow focus overlooks the broader spectrum of visual concepts, which includes not only concrete entities like individuals but also general contexts such as event scenes (Fig. \ref{fig:figure1}(a)). Critically, current benchmarks fail to account for the coupling of these concepts, as a single image inherently contains semantically entangled elements. This raises a question: \textit{when forgetting an entity, does it affect other concepts within the same image (such as event scenes)?} Such limitations lead to an insufficient evaluation of an MU method's true efficacy, which motivates our work to develop a more comprehensive benchmark.

We categorize multimodal unlearning targets into two types based on the scope of visual concepts: \textcolor{Highlight}{Person (Fig. \ref{fig:figure1}(b))} and \textcolor{darkpurple}{Event (Fig. \ref{fig:figure1}(c))}. This distinction is motivated by real-world challenges that raise significant privacy and safety concerns in multimodal learning. Specifically, the Person category relates to protecting personal privacy and intellectual property, while the Event category involves removing potentially harmful or illegal content, such as scenes depicting fake news. Accordingly, we introduce a new benchmark named \textbf{PEBench}, which is explicitly designed to assess the unlearning of both \textbf{P}erson and \textbf{E}vent concepts.

PEBench is constructed using synthetic data, offering two primary advantages:

(1) \textbf{Establish a reliable upper bound for unlearning performance.} While retraining a model from scratch after removing target data represents the gold standard for exact unlearning~\cite{wang2023machine,guo2023verifying}, this approach is computationally infeasible for LLMs. TOFU~\cite{maini2024tofu} addresses this issue by introducing fictitious data that was never seen during pretraining, allowing fine-tuned LLMs to approximate the behavior of retraining. Specifically, TOFU features 200 fictitious author profiles, each described by attributes such as name, birthplace, parents' names, and occupations, including 4,000 question-answer pairs for evaluating LLMs' unlearning. CLEAR~\cite{dontsov2024clear} extends TOFU to the MLLMs by generating consistent synthetic images associated with the TOFU authors over 200 fictitious identities. Following this paradigm, PEBench also comprises 200 fictitious individuals, but significantly expands its complexity by linking each individual to 40 distinct event scenes, creating a dataset of 8,000 images for more nuanced evaluations.


\textbf{(2) Ensure controllable data quality for satisfying both generality and scope evaluation.} Generality assesses whether a model can forget a concept across diverse samples~\cite{liu2025rethinking}, whereas scope measures if only the intended target is forgotten while preserving unrelated information~\cite{hase2021language, cohen2024evaluating}. PEBench is therefore synthetically generated to ensure both high intra-target consistency (for evaluating generality) and deliberate inter-concept coupling (for evaluating scope). This design facilitates a reliable and fine-grained assessment of unlearning efficacy and its potential side effects.


We benchmark 5 MU methods, providing new insights into their strengths and weaknesses in unlearning personal entities and events, as well as their underlying mechanisms. Moreover, we demonstrate that existing methods struggle to unlearn both concepts simultaneously due to their semantic entanglement. To address this, we introduce a simple yet effective method to mitigate this challenge. We believe our benchmark will significantly contribute to the future development of MU for MLLMs. Key contributions are summarized as follows:

\begin{itemize}
    \item We introduce PEBench, a novel benchmark for evaluating Machine Unlearning (MU) in Multimodal Large Language Models (MLLMs). Its synthetic dataset of 8,000 images, featuring 200 fictitious personal entities across 40 event scenes, is specifically designed to address the conceptual limitations of existing benchmarks.

    \item Facilitates a fine-grained evaluation of unlearning on two distinct conceptual targets: private information (Person) and general scenes (Event). By enforcing intra-target visual consistency and inter-concept coupling, its synthetic design enables the controlled assessment of key metrics, including the \textbf{upper bound of unlearning efficacy} and the \textbf{scope of unlearning}.

    \item We provide a comprehensive benchmark of 5 different MU methods, revealing their respective strengths and weaknesses for person and event unlearning. Furthermore, we propose and validate an effective method to address the more complex challenge of simultaneously unlearning both targets.

\end{itemize}

\section{Related Work}

\subsection{Machine Unlearning}

Motivated by growing privacy and security concerns~\cite{hoofnagle2019european,zhang2024forgetting}, machine unlearning (MU) was introduced in~\cite{cao2015towards} to enable the removal of toxic or biased content from machine learning models. Recent work has increasingly focused on unlearning in large language models (LLMs)~\cite{nguyen2022survey,li2024wmdp}, where the scale and complexity exacerbate the risk of memorizing sensitive information. Representative methods include gradient-based approaches like Gradient Ascent (GA)~\cite{yao2023large} and Gradient Difference (GD)~\cite{liu2022continual}. These techniques adjust model parameters to induce mispredictions on the data intended for forgetting (the forget set)~\cite{jang2022knowledge}. However, a significant drawback of these methods is the risk of catastrophic forgetting~\cite{liu2024rethinking}, where the model's performance on unrelated, retained data is excessively degraded. To mitigate this issue, one approach is to use KL-divergence regularization~\cite{yao2024machine} to maintain the model's behavior on the retain set. Another prominent direction is preference-based optimization. This includes reinforcement learning frameworks that use task-specific reward functions~\cite{lu2022quark}, as well as simpler alignment techniques like Preference Optimization (PO)~\cite{maini2024tofu} and Direct Preference Optimization (DPO)~\cite{rafailov2023direct}, which only require positive and negative response pairs. Given that both LLMs and multimodal large language models (MLLMs) adopt an auto-regressive transformer-based architecture, these unlearning approaches can be extended to the multimodal setting~\cite{maini2024tofu}. In this paper, we systematically evaluate these methods in MLLMs.

\begin{table*}[!t]
    \centering
    \small
    \caption{
    \small{The comparison between PEBench and other MU benchmarks. 
    }
    \vspace{-2mm}
    }
   
    \resizebox{\linewidth}{!}{
    \begin{tabular}{lllllll} 
        \toprule
        \textbf{Benchmarks} & \textbf{Image type} & \textbf{Images source} & \textbf{Image Number}& \textbf{Multimodel QA pairs}&\textbf{Image Number pre Concept}&\textbf{Visual Concept}
        \\
        \midrule
         
          MMUBench~\cite{li2024single} &Real World &Mike&1,000 &1,000  & 50 / Entity & 20 Entities \\
          CLEAR~\cite{dontsov2024clear} &Synthetic  &StyleGAN2&3,770 &4,000 & 15-20 / Entity & 200 Entities \\
          MLLMU~\cite{liu2024protecting}&Synthetic  &StyleGAN2&1,153
&10,377  & 2 / Entity & 500 Entities  \\
          FIUBench~\cite{ma2024benchmarking} &Synthetic  &StyleGAN2&400 &8,000
  & 1 / Entity & 400 Entities  \\
  \midrule
          \textbf{PEBench} &Synthetic  &FLUX&8,000 &16,000 & 40 / Entity, 200 / Event & 200 Entities \& 40 Events   \\
         \bottomrule
    \end{tabular}
    }
     
    \label{table: dataset_comparison}
\end{table*}

\subsection{Unlearning Benchmarks}
\label{Unlearning Benchmarks}

Standardized benchmarks are critical for the rigorous evaluation of machine unlearning (MU) methods~\cite{wang2025evaluation}. In the LLM domain, various datasets have been used to assess the removal of harmful content~\cite{lu2022quark}, personal identifying information~\cite{jin2024rwku,patil2023can}, and copyrighted material~\cite{eldan2023s}. For MLLMs, the first dedicated benchmark was MMUBench ~\cite{li2024single}, built upon the MIKE dataset~\cite{li2024mike}. Its reliance on real-world data presents an inherent challenge: the 'gold standard' for unlearning, retraining a model from scratch without the target data, is computationally infeasible at scale, limiting definitive evaluation. To address this, recent efforts have increasingly adopted synthetic data as a practical alternative for simulating the retraining process. TOFU~\cite{maini2024tofu} introduced 200 fictitious author profiles and 4,000 question-answer pairs to enable controlled and reproducible MU evaluations in LLMs. Building on this paradigm, CLEAR~\cite{dontsov2024clear} extended MU assessment to multimodal settings by generating synthetic facial images for each TOFU identity. Similarly, MLLMU~\cite{liu2024protecting} and FIUBench~\cite{ma2024benchmarking} employ synthetic image generation pipelines such as StyleGAN2 and ThisPersonDoesNotExist to construct high-fidelity, privacy-preserving datasets for MLLM unlearning. Despite these advances, most existing benchmarks focus exclusively on entity-level unlearning and overlook broader visual concepts such as scene or event semantics. To bridge this gap, we propose PEBench, a new benchmark that evaluates MU in both personal identity and event-centric contexts. As summarized in Table~\ref{table: dataset_comparison}, PEBench provides a more scalable and comprehensive framework for assessing MU performance in realistic multimodal scenarios.


\section{PEBench}

\begin{figure*}[t] \centering
\centering
  \resizebox{1\linewidth}{!} {
    \includegraphics{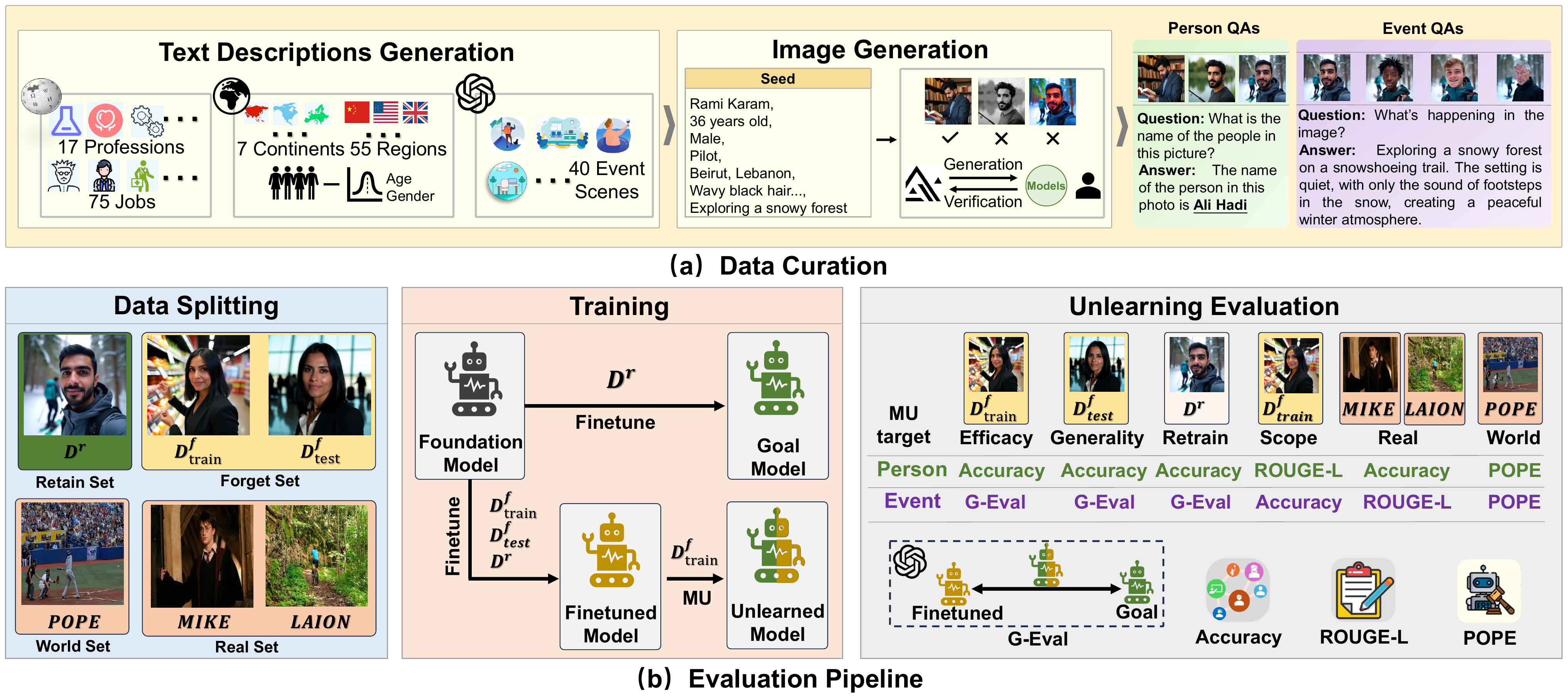}
  }
  \vspace{-15pt}  
  \caption{Overview of PEBench. (a) Data Curation: We construct a synthetic dataset of 200 individuals across diverse occupations, regions, and demographics, each paired with 40 event scenes. Images are generated and filtered for both person and event consistency. (b) Evaluation Pipeline: The dataset is split into Forget, Retain, Real, and World sets. A foundation model is fine-tuned on the Retain set to obtain a Goal Model and train the Unlearned Model on PEBench. PEBench evaluates unlearning across six dimensions using metrics such as Accuracy, ROUGE-L, POPE, and G-Eval.}
    \label{fig:main}
\end{figure*}

\subsection{Data Curation}
As illustrated in Fig.~\ref{fig:main}(a), this process involves two steps: generating text descriptions for person-event pairs and then synthesizing the corresponding images.

\noindent \textbf{Text Description Generation.} For persons, we define key attributes (such as profession, age, gender, and birthplace) which serve as prompts for GPT-4~\cite{achiam2023gpt} to generate detailed character profiles. Professions span 17 major categories and 75 specific occupations (sourced from Wikipedia\footnote{https://en.wikipedia.org/wiki/Lists\_of\_occupations}), while birthplaces are sampled from 7 continents and 55 regions. These attributes are randomly combined to produce diverse, fictitious individuals with distinct names and appearances. Consistent with prior benchmarks like TOFU~\cite{maini2024tofu} and CLEAR~\cite{dontsov2024clear}, which also adopt 200 synthetic identities, we generate 200 virtual personal entities striking a balance between diversity and manageability.

\begin{tcolorbox}[title=GPT-4 Prompting Strategy for Character Descriptions, colback=gray!20, colframe=gray!75, rounded corners, sharp corners=northeast, sharp corners=southwest]
\textbf{Person Prompt:}
I want to create a fictional character with the following attributes:\\
Job: \{\}\\
Birthplace: \{\}\\
Age: \{\}\\
Gender: \{\}\\
generate a character's appearance and name:
\vspace{-2mm}
\end{tcolorbox}

For the event, GPT-4 is prompted to generate 40 thematically diverse and richly detailed scene descriptions. In contrast to basic image captions, these scenes are intentionally designed to convey vivid contextual and semantic information, which enhances the realism and challenge of the machine unlearning tasks. To ensure the reliability of the generated events, all descriptions undergo manual review to confirm their semantic coherence and alignment with the intended themes.

\begin{tcolorbox}[title=GPT-4 Prompting Strategy for Event Descriptions, colback=gray!20, colframe=gray!75, rounded corners, sharp corners=northeast, sharp corners=southwest]
\textbf{Events Prompt:} Please generate 40 event scenarios, each with a distinct theme and a detailed description of its features. The description of each event should include the following:\\
Theme: The core content or purpose of the event.\\
Features: Specific details of the event, such as participants, environment, atmosphere, location, etc., ensuring the description is vivid, concrete, and visually rich.
\end{tcolorbox}

\noindent \textbf{Image Generation.}
A primary challenge in image generation is maintaining the consistent appearance of an individual across multiple events while also preserving stylistic coherence across different characters. Standard text-to-image models such as Stable Diffusion~\cite{rombach2022high} often have difficulty with both identity and scene consistency.

While methods like IPAdapter~\cite{ye2023ip} and PhotoMaker~\cite{li2024photomaker} have been used to improve image consistency, they frequently fail to generate recognizable character appearances and can lack realism (Fig.~\ref{fig:photomaker}). To address these limitations, we employ a two-step strategy: \textit{generate-then-filter}. The initial step focuses on guiding the image generation toward high visual quality, while the second step enforces consistency and fidelity via post-generation filtering.

We use Flux\footnote{\url{https://github.com/black-forest-labs/flux}} as our image generator, as it is particularly effective at generating images that are both realistic and stylistically coherent~\cite{flux2024}. The input prompt is carefully constructed to control the identity, appearance, and event context, and is formatted as follows:

\begin{tcolorbox}[title=Flux Prompting Strategy for Images Generation, colback=gray!20, colframe=gray!75, rounded corners, sharp corners=northeast, sharp corners=southwest]

\textbf{Images Generation Prompt:} A photo of a \textless Job\textgreater{} \textless Age\textgreater{}-year-old \textless Gender\textgreater{} from \textless Birthplace\textgreater{}, named \textless Name\textgreater{}, with \textless character’s appearance\textgreater{}.:\\
\textless events\textgreater{}\\
Generate this photo with the Person ID: \textless Person ID\textgreater{} and with file name IMG\_6105.CR2.
\end{tcolorbox}

\begin{figure*}[ht]
\centering
  \resizebox{1\linewidth}{!} {
    \includegraphics{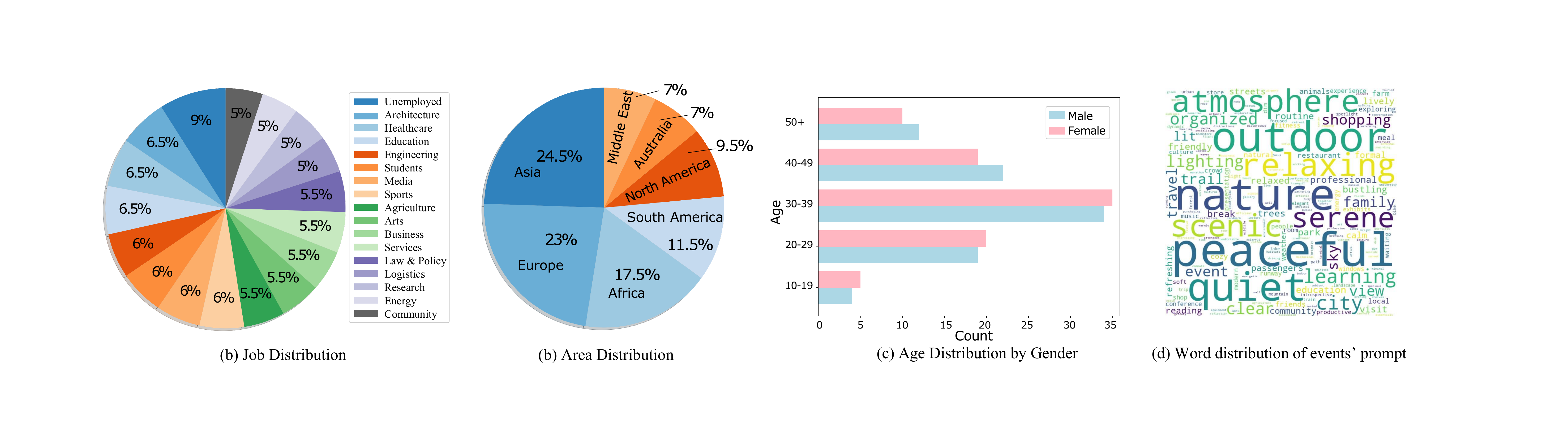}
  }
  \vspace{-15pt}  
  \caption{(a) Age distribution by gender shows a balanced representation across age groups and between male and female individuals. (b) Area distribution demonstrates geographic diversity, with individuals from various continents, ensuring a globally representative dataset. (c) Job distribution displays a wide range of professions, covering multiple fields such as healthcare, education, arts, business, and technology, highlighting the dataset's comprehensive coverage of occupational diversity.}
    \label{fig:analysis}
\end{figure*}

\begin{figure}[tb] \centering
    \includegraphics[width=0.45\textwidth]{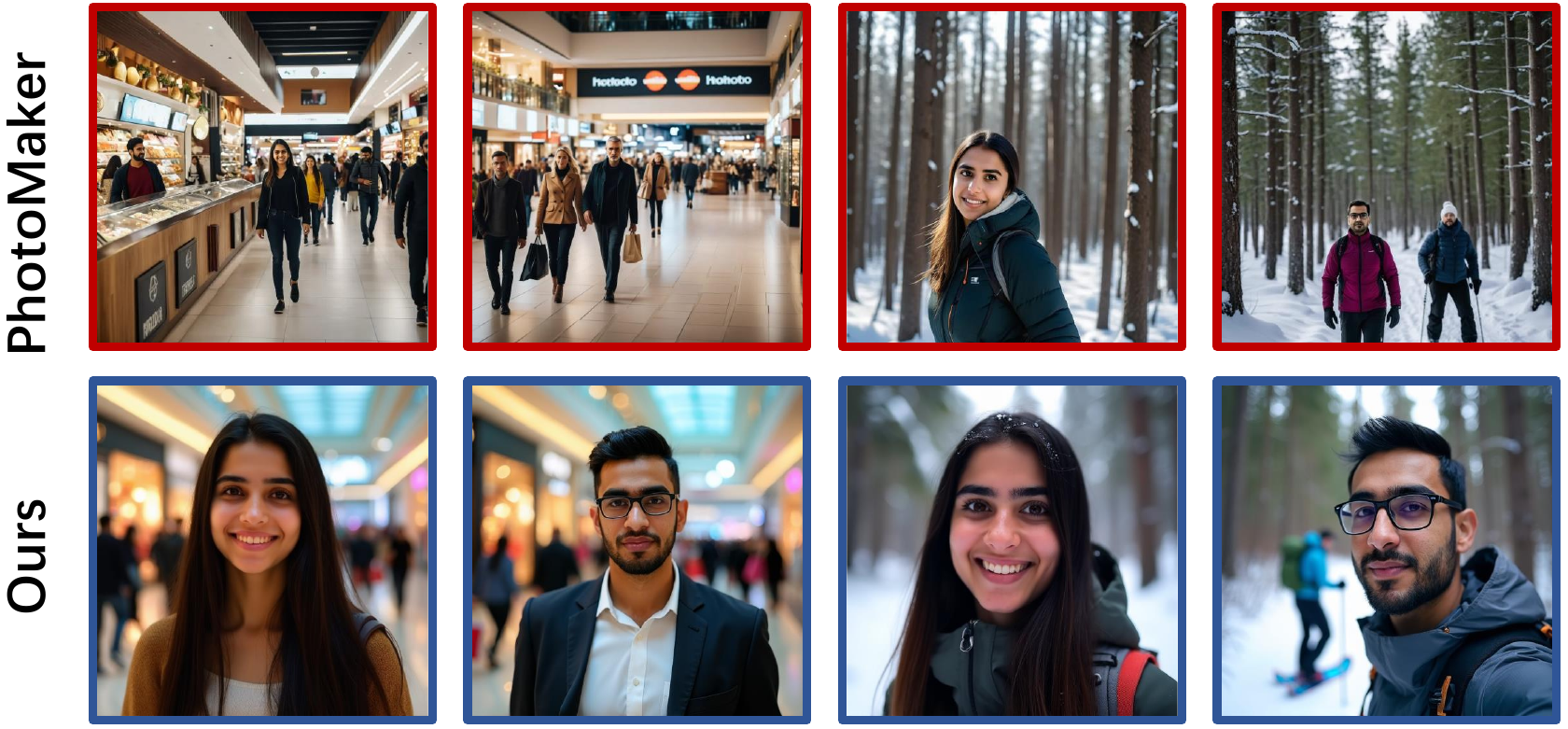}
    \vspace{-5pt}  
\caption{Comparison of image generation results between our method and PhotoMaker. The images generated by our method demonstrate consistency in both character appearance and scene setting across different contexts.} 

\label{fig:photomaker} 
\end{figure}

\begin{figure}[tb] \centering
    \includegraphics[width=0.45\textwidth]{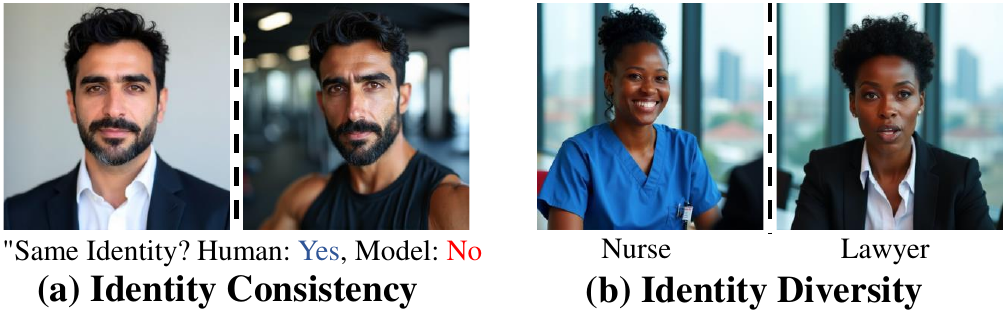}
    \vspace{-5pt}  
\caption{(a) Identity consistency is effectively checked and ensured by the facial recognition model, even for samples that may be difficult for humans to distinguish. (b) A person's professional attributes influence the appearance characteristics of the virtual character.} \label{fig:job} 
\end{figure}

A multi-stage filtering process is subsequently applied to ensure consistency in both character appearance and scene style. This process involves three key stages:

\noindent \textbf{i) Identity Consistency.}
We use FaceNet~\cite{schroff2015facenet} to verify that each virtual character maintains a consistent facial appearance across different scenes. Images with an L2 distance exceeding 0.8 between facial embeddings are considered inconsistent and are regenerated (Fig.~\ref{fig:job}(a)).

\noindent \textbf{ii) Background Consistency.}
To maintain scene coherence, we compute pairwise CLIP~\cite{Radford2021LearningTV} feature similarities among images of the same event. Images with cosine similarity below 0.3 are filtered out.

\noindent \textbf{iii) Image Quality.}
To ensure high data quality, we adopt image quality assessment (IQA) techniques inspired by established image and video benchmarks~\cite{huang2024vbench}. This filtering process evaluates images across three dimensions:

\noindent \textbf{(1) Subject Consistency.}
DINO~\cite{caron2021emerging} features are extracted to verify appearance consistency. Images with cosine similarity below 0.85 are discarded.

\noindent \textbf{(2) Distortion Detection.}
The MUSIQ~\cite{ke2021musiq}, trained on SPAQ~\cite{fang2020perceptual}, is used to identify artifacts such as blur or noise. Images scoring below 50 are removed.

\noindent \textbf{(3) Aesthetic Quality.}
Visual appeal, including composition, color harmony, and photo-realism, is evaluated using the LAION aesthetic predictor~\cite{laion2022aesthetic}. Images with a score below 6.0 are excluded.

For each prompt, five image samples are generated by varying the guidance scale and seed. If none meet the above criteria, a new batch is synthesized until at least one valid sample is retained.

\subsection{Quality Assurance}

\noindent \textbf{Data Statistics.}
The dataset was carefully designed to ensure diversity and representativeness by balancing key demographic attributes. As shown in Fig.\ref{fig:analysis}, it features a well-distributed composition across professions, geographic regions, age groups, and gender. While certain attributes (e.g., gender, age, or region) may overlap across individuals, their differences are visually reflected in each character’s appearance. For example, professional roles lead to distinct variations in clothing styles and contextually appropriate accessories, as illustrated in Fig.\ref{fig:job}(b).

Additionally, to support scene variability, our prompt suite includes a wide range of contextually rich event descriptions. A word cloud summarizing the semantic distribution of event-related prompts is provided in Fig.~\ref{fig:analysis}(d).

\noindent \textbf{Human Verification}  
To ensure the generated images maintain high quality and align with the data statistics, the dataset was manually evaluated by two authors of this paper and two volunteers with undergraduate degrees. The verification process consists of two aspects: appearance consistency and image quality:

\textbf{i) Appearance Consistency.} The appearance of all 200 individuals was manually validated to ensure alignment with their respective prompt descriptions. An image was deemed valid only after all four reviewers confirmed its consistency across key attributes, including region, age, profession, and physical features. If an image was not validated, it was regenerated with an adjusted random seed.

\textbf{ii) Image Quality Check.} To assess the effectiveness of automatic filtering, a comparative evaluation was performed on 10\% of the original samples, which found that 98\% of the images identified as low-quality by human annotators were also flagged by the automated filters. This demonstrates a strong alignment between human judgment and model-based assessments. The final dataset was first filtered using the model-based quality assessment, with the results then being double-checked by human annotators to guarantee 100\% quality compliance.

\subsection{Evaluation Pipeline}

As shown in Fig.~\ref{fig:main} (b), we establish a structured evaluation pipeline composed of three components: dataset splitting (Sec~\ref{sec:Datasets_Splits}), model training (Sec~\ref{sec:Training}), and performance evaluation (Sec~\ref{sec:metri}).

\subsubsection{Datasets Splitting}
\label{sec:Datasets_Splits}
Our framework incorporates four distinct datasets: the Forget Set, Retain Set, Real Set, and World Set. Detailed descriptions of each are provided below.

\textbf{Forget Set}. 
We define two forget targets, \textit{person} and \textit{event}, evaluated independently. For each target, we randomly sample 5\%, 10\%, or 15\% of the corresponding instances from the full dataset. All images linked to these selected individuals or events form the Forget Set, which is evenly split into $\mathcal{D}^f_{\text{train}}$ for unlearning training and $\mathcal{D}^f_{\text{test}}$ for generalization evaluation.

\textbf{Retain Set}.
The Retain Set $\mathcal{D}^r = \mathcal{D} \setminus \mathcal{D}^f$ comprises the remaining data not included in the Forget Set. It is used to evaluate the model’s ability to retain non-targeted knowledge post-unlearning.

\textbf{Real Set}.
The Real Set consists of real-world images that are semantically aligned with concepts in both the Forget and Retain Sets. For person unlearning, it includes public figures from the MIKE dataset~\cite{li2024mike}, such as “Trump”. For event unlearning, we retrieve semantically similar images from the LAION-5B~\cite{laion2022aesthetic} based on event descriptions. Each real-world target type (person or event) is represented by 200 images.

\textbf{World Set}. To evaluate whether general factual knowledge is preserved, we include a World Set sampled from the POPE benchmark~\cite{Li-hallucination-2023}.

\subsubsection{Training}
\label{sec:Training}
The gold standard for machine unlearning (MU) is retraining a model from scratch without the forget set~\cite{liu2024rethinking}. To simulate this ideal scenario using synthetic data, we partition the dataset $\mathcal{D}$ into a Forget Set $\mathcal{D}^f$ and a Retain Set $\mathcal{D}^r$, and train three models:

\textbf{Goal Model.} This model is fine-tuned exclusively on $\mathcal{D}^r$, serving as the ideal reference where all sensitive data has been excluded. Since the dataset comprises synthetic content with fictitious names and events, it can be treated as unseen data for MLLMs. The Goal Model provides an \textbf{upper bound} for evaluating unlearning success.

\textbf{Finetuned Model.} This model is trained on the full dataset $\mathcal{D}$ (including $\mathcal{D}^f$) and represents the \textbf{baseline} before unlearning is applied.

\textbf{Unlearned Model.} Starting from the finetuned model, unlearning algorithms are applied using $\mathcal{D}^f_{\text{train}}$ to remove the influence of the forget set. The resulting model is evaluated against both the goal and finetuned models to assess unlearning efficacy and model utility preservation.

\subsubsection{Unlearning Evaluation}
\label{sec:metri}

We introduce 6 evaluation metrics within the PEBench framework, covering both forgetting effectiveness and model utility:


{\textbf{Efficacy}} measures how effectively the model forgets the target concepts in $\mathcal{D}^f_{\text{train}}$. For person unlearning, it evaluates name recognition accuracy. For event unlearning, it uses GPT-4 Evaluation to assess the model’s deviation from event-specific knowledge (detailed later in this section).

{\textbf{Generality}} evaluates the model’s ability to generalize forgetting beyond memorized samples. This is done by testing on $\mathcal{D}^f_{\text{test}}$, which contains novel instances of the same individuals or events not seen during unlearning.

{\textbf{Retain}} assesses the model’s performance on the Retain Set $\mathcal{D}^r$ to ensure that non-targeted knowledge remains unaffected, thereby preserving the model’s overall utility.

{\textbf{Scope}} measures whether forgetting one concept unintentionally affects related concepts in the same image. When unlearning persons, we compute the ROUGE-L score between the model’s generated scene descriptions and ground truth on $\mathcal{D}^f_{\text{train}}$. Conversely, for event unlearning, we evaluate the accuracy of person recognition.

{\textbf{Real}} evaluates the model’s ability to retain real-world knowledge relevant to PEBench. For person unlearning, we assess name recognition accuracy on the person portion of the Real Set. For event unlearning, we compute the ROUGE-L similarity between the outputs of the unlearned and finetuned models, where lower similarity indicates greater degradation in model utility.

{\textbf{World Fact}} assesses whether unlearning degrades factual knowledge. We use the POPE benchmark~\cite{Li-hallucination-2023} to measure the preservation of general world knowledge.


\noindent \textbf{GPT-4 Evaluation (G-Eval).}
In the context of event unlearning, the outcome of forgetting is inherently non-deterministic. Therefore, instead of using the original ground truth as the sole reference, we adopt the output of a predefined goal model as the reference for successful unlearning. Conversely, the fine-tuned model (before unlearning) serves as the baseline.

To assess the similarity between the unlearned model’s output and both the goal and fine-tuned models, we employ the LLM-as-a-judge framework~\cite{zheng2023judging}, which has demonstrated superior alignment and semantic evaluation capabilities compared to traditional surface-level metrics~\cite{liu2022revisiting, wang2023chatgpt}. Specifically, GPT-4o-mini (balances the cost and quality of evaluation~\cite{singh2024scidqa}) is prompted to assign a similarity score ranging from 0 to 1: a score of 0 indicates complete similarity with the fine-tuned model (i.e., no forgetting), while a score of 1 indicates complete similarity with the goal model (i.e., full forgetting). For consistency with other metrics in our benchmark, the G-Eval scores are scaled by a factor of 100.

\vspace{2mm}
\begin{tcolorbox}[title= Instructions for G-Eval, colback=gray!20, colframe=gray!75, rounded corners, sharp corners=northeast, sharp corners=southwest]
\vspace{2mm}
\textbf{Images Generation Prompt:} Evaluate the effectiveness of unlearning by comparing the test sentence (produced by the unlearned model) against two references: the goal sentence (from the goal model) and the original sentence (from the fine-tuned model before unlearning).

Assign a similarity score between \textbf{0 and 1}, where:

\begin{itemize}
\item A score closer to \textbf{1} indicates the test sentence is more similar to the goal sentence, suggesting effective unlearning (i.e., the target knowledge has been successfully forgotten).
\item A score closer to \textbf{0} indicates the test sentence is more similar to the original sentence, suggesting ineffective unlearning (i.e., the target knowledge has not been removed).
\end{itemize}
\end{tcolorbox}
\vspace{2mm}

\section{Experiment}
\begin{table*}[t]
\centering

\vspace{2mm}
\centering
\caption{Hyperparameter settings for fine-tuning and unlearning. 
    }
\begin{tabular}{l|c|c|c|c|c|c|lllll}
\toprule

\multirow{2}{*}{MLLMs} &\multirow{2}{*}{Epoch} &Batch&Gradient &\multirow{2}{*}{Optimizer}& \multirow{2}{*}{LoRA}&Finetune&\multicolumn{5}{c}{ Unlearning Methods Learning Rate}  \\
\cline{8-12}
&  & Size &Accumulation& &  &Learning Rate&PO&GA&GD&KL&DPO \\
 
 \midrule
\multirow{1}{*}{LLaVA-1.5-7B}
 & 5 &4 &16 & AdamW & True & 2e-5 & 3e-4& 2e-5& 2e-5& 1e-4& 2e-5\\

\multirow{1}{*}{Llama-3.2-Vision-11B} 
 & 5 & 2& 16 & AdamW & True & 2e-5 & 3e-4& 2e-5& 2e-5& 1e-4 & 2e-5\\
 \bottomrule
\end{tabular}

\label{tab:appendix-param}
\end{table*}
\subsection{Unlearning Methods}

We assess the 5 recently proposed MU methods on PEBench. Detailed descriptions for each method follow:

\noindent
\textbf{Gradient Ascent (GA)~\cite{yao2023large}}: is a fundamentally straightforward approach, reducing the likelihood of correct predictions on the forget set. It updates the model parameters $w$ by maximizing the likelihood of mis-prediction for the samples within the forget set $\mathcal{D}^f$. Given a sample $\textbf{x} \in \mathcal{D}^f$, the loss can be denoted by: 
\begin{equation}
L(\mathcal{D}^f, w) =  \frac{1}{\vert \mathcal{D}^f \vert} \sum_{x \in \mathcal{D}^f}\ell(\textbf{x}, w). 
\end{equation}

\noindent
\textbf{Preference Optimization (PO)~\cite{maini2024tofu}}: This approach guides the model to align with newly generated responses such as “I do not know the answer” and its variants for questions related to the forget set $\mathcal{D}^f$. Simultaneously, it incorporates a retain set term to ensure the model's predictions for the retained set remain unaffected. The objective function is defined as:

\begin{equation}
L_{\text{idk}} = L(\mathcal{D}^r, w) + L(\mathcal{D}^f_\text{idk}, w).
\end{equation}

\noindent
\textbf{Gradient Difference (GD)~\cite{liu2022continual}}, 
This method extends gradient ascent by simultaneously focusing on forgetting the samples in the forget set \(\mathcal{D}^f\) and preserving performance on the retain set \(\mathcal{D}^r\). The objective is to balance increasing the loss for the forget set and minimizing the impact on the retained set. The resulting loss function to be minimized is expressed as:

\begin{equation}
    L_{\text{diff}} = - L(\mathcal{D}^f, w) + L(\mathcal{D}^r, w).
\end{equation}

\noindent  
\textbf{KL~\cite{yao2024machine}}:  
This method extends Gradient Ascent by incorporating an additional objective to minimize the Kullback-Leibler (KL) divergence between the predictions of the original model \(M_{\text{ori}}\) and the newly trained model \(M_{\text{new}}\) on the retain set \(\mathcal{D}^r\). The KL divergence loss is defined as:

\begin{gather}
L_{KL} = 
\frac{1}{\vert \mathcal{D}^r \vert} \sum_{s \in \mathcal{D}^r} \frac{1}{\vert s \vert} \sum_{i = 2}^{\vert s \vert} \text{KL}\left(M_{\text{ori}}(s_{<i}) \big\Vert M_{\text{new}}(s_{<i})\right). 
\end{gather}

\noindent  
The overall objective function combines the Gradient Ascent loss on the forget set and the KL divergence loss:

\begin{equation}
L_{\text{total}} = - L(\mathcal{D}^f, w) + L_{KL}.
\end{equation}

\noindent  
\textbf{Direct Preference Optimization (DPO)~\cite{rafailov2024direct}}:  
DPO directly optimizes language models to align with human preferences without the need for explicit reward modeling or reinforcement learning. For unlearning, this approach is framed as a preference optimization problem, where the preference is shifted towards outputs that relabel or neutralize unwanted data. This ensures the model effectively forgets targeted information while aligning with desired outputs. The loss function is defined as:

\begin{equation}
\begin{split}
L_{DPO}(\pi_{\theta}, \pi_{ref}) =  
-\mathop{\mathbb{E}_{\textbf{x}, y \in \mathcal{D}^f}}_{y' \in \mathcal{D}^f_\text{idk}} \Big[  \\
    \log \sigma\Big(\beta\log\frac{\pi_{\theta}(y'|\textbf{x})}{\pi_{ref}(y'|\textbf{x})} - 
    \beta\log\frac{\pi_{\theta}(y|\textbf{x})}{\pi_{ref}(y|\textbf{x})}\Big) \Big],
\end{split}
\end{equation}

\noindent
where \(L_{DPO}\) is the preference optimization loss, \(\pi_{\theta}\) and \(\pi_{ref}\) represent the unlearning target model and the reference model trained on \(\mathcal{D}^f_\text{idk}\), respectively. \(\sigma\) is the logistic function, and \(\beta\) is the DPO scaling coefficient.

The total objective function combines task performance and unlearning effectiveness and is defined as:

\begin{equation}
L = \lambda_1 L(\mathcal{D}^f_\text{idk}, \theta) + \lambda_2 L_{DPO}(\pi_{\theta}, \pi_{ref}),
\end{equation}

where \(\lambda_1\) and \(\lambda_2\) are weighting values that balance task performance and the unlearning process.

\subsection{Experiment setup}

All experiments, including both fine-tuning and unlearning, are conducted using two NVIDIA A100 GPUs (80GB). We evaluate on two representative MLLMs: LLaVA-1.5-7B~\cite{liu2023llava} and LLaMA-3.2-Vision-11B~\cite{meta2024llama}. The corresponding hyperparameter configurations are provided in Table~\ref{tab:appendix-param}.

\begin{table*}[t]
\centering
\vspace{2mm}
\caption{Performance overview of different MU methods evaluated on PEBench. The performance metrics include Efficacy, Generality, Retain, Real, and World Fact. A higher score represents better performance. \colorbox{lightgray}{Finetune} represents the baseline unlearning capability (lower bound for unlearning), and \colorbox{pink}{Goal} represents the ideal unlearning model (upper bound). Highest and second best are highlighted in \textbf{bold} and \underline{underline}, respectively.}

\resizebox{\textwidth}{!}{%
\begin{tabular}{c|cccccc|cccccc}
\toprule
\multirow{3}{*}{\textbf{Method}} & \multicolumn{6}{c|}{\textbf{Person Unlearning}} & \multicolumn{6}{c}{\textbf{Event Unlearning}} \\
\cline{2-13}
& \textbf{Efficacy}  & \textbf{Generality}  & \textbf{Retain} & \textbf{Scope} & \textbf{Real}  & \textbf{World Fact}  &\textbf{Efficacy}  & \textbf{Generality}  & \textbf{Retain} & \textbf{Scope} & \textbf{Real}  & \textbf{World Fact} \\

& \small{Precision}  & \small{Precision}  & \small{Precision} &  \small{ROUGE-L} & \small{Precision}  & \small{POPE}   &\small{G-Eval}  & \small{G-Eval}  & \small{ROUGE-L} & \small{Precision} & \small{ROUGE-L}  & \small{POPE} \\
\midrule
\multicolumn{13}{c}{\textbf{LLaVA-1.5-7B (5\% Forget)}} \\
\midrule
\rowcolor{lightgray}  
Finetune&2.0&1.5&94.8&81.1&67.7&82.3&18.0&21.3&84.4&92.2&39.6&82.3\\

PO~\cite{maini2024tofu}         &88.4&89.1&\underline{18.0}&\underline{51.8}&14.9&\underline{81.9}&43.9&46.5&\textbf{50.8}&\underline{80.2}&\underline{26.8}&81.0\\
GA~\cite{yao2023large}          &\textbf{100.0}&\textbf{100.0}&3.3&8.8&6.9&80.2&\underline{46.4}&\underline{46.8}&22.3&28.8&6.1&80.0\\
GD~\cite{liu2022continual}      &\textbf{100.0}&99.0&9.3&33.6&\underline{21.7}&78.0&44.6&\textbf{47.0}&27.3&62.5&\textbf{27.5}&\underline{81.6}\\
KL~\cite{yao2024machine}        &\underline{99.5}&\underline{99.5}&\textbf{18.2}&\textbf{57.0}&\textbf{24.6}&\textbf{83.1}&\textbf{48.9}&\textbf{47.0}&\underline{31.6}&\textbf{82.0}&{26.0}&\textbf{81.7} \\
DPO~\cite{rafailov2024direct}   &\textbf{100.0}&\textbf{\textbf{100.0}}&5.8&20.1&3.6&80.2&43.2&44.2&24.1&51.3&11.3&81.1\\
\rowcolor{pink}  
Goal
&100.0&100.0&95.0 &83.8&71.8&82.8&91.9&86.8&85.6&94.5&40.3&82.8\\
\midrule
\multicolumn{13}{c}{\textbf{LLaVA-1.5-7B (10\% Forget)}} \\
\midrule
\rowcolor{lightgray}  
Finetune&5.6&6.3&95.5&81.5&67.7&82.3&18.1&17.6&83.2&94.0&39.6 &82.3\\

PO~\cite{maini2024tofu}         &82.5&76.3&\underline{29.5}&\textbf{50.1}&10.1&{81.8}&30.2&30.9&\textbf{35.8}&56.0&\underline{18.6}&\textbf{82.6}\\
GA~\cite{yao2023large}          &\textbf{100.0}&\underline{99.5}&2.2&3.5&\underline{21.4}&80.8&\underline{46.9}&\underline{47.7}&7.1&37.7&4.4&80.3\\
GD~\cite{liu2022continual}      &\textbf{100.0}&\textbf{100.0}&10.3&35.5&\textbf{54.0}&\underline{82.4}&41.6&40.2&12.7&\underline{65.7}&6.6&80.9\\
KL~\cite{yao2024machine}        &\underline{96.0}&96.4&\textbf{35.5}&\underline{38.8} &12.1&\textbf{82.7}&\textbf{48.6}&\textbf{48.1}&\underline{33.0}&\textbf{82.2}&\textbf{38.0}&\underline{82.2} \\
DPO~\cite{rafailov2024direct}   &\textbf{100.0}&\textbf{100.0}&3.3&3.7 &2.4&72.7&35.7&32.4&9.8&41.0&0.4&80.8\\
\rowcolor{pink}  
Goal
&100.0&100.0&96.5&82.5&71.1&83.6&89.3&91.6&86.8&94.6&40.1&83.6\\
\midrule
\multicolumn{13}{c}{\textbf{LLaVA-1.5-7B (15\% Forget)}} \\
\midrule
\rowcolor{lightgray}  
Finetune&4.1&6.7&95.0&79.3&67.7&82.3&17.6&19.3&82.4&93.0& 39.6&82.3\\

PO~\cite{maini2024tofu}         &75.0&80.0&22.2&\textbf{56.6}&28.2&\underline{82.0}&40.1&\textbf{40.9}&\textbf{36.5}&\underline{69.5}&\underline{12.6}&\textbf{82.3}\\
GA~\cite{yao2023large}          &\textbf{100.0}&\textbf{100.0}&\underline{23.7}&0.5&6.1&79.7&35.4&31.8&3.1&6.3&0.1&79.5\\
GD~\cite{liu2022continual}      &\textbf{100.0}&\textbf{100.0}&10.5&41.7&\textbf{61.2}&80.4&36.8&34.7&12.0&44.5&6.7&81.0\\
KL~\cite{yao2024machine}        &\underline{92.5}&\underline{92.4}&\textbf{34.3}&\underline{55.2}&\underline{48.8}&\textbf{82.2}&\textbf{42.9}&\underline{40.7}&\underline{29.1}&\textbf{90.7}&\textbf{27.8}&\underline{82.1}\\
DPO~\cite{rafailov2024direct}   &\textbf{100.0}&\textbf{100.0}&13.3&6.1&6.9&71.4&\underline{40.9}&40.5&3.9&22.0&5.6&81.4\\
\rowcolor{pink}  
Goal
&100.0&100.0&96.3&82.4&68.5&83.5&88.5&90.8&85.4&94.6&42.4&83.5\\
\midrule
\multicolumn{13}{c}{\textbf{LLama-3.2-Vision-Instruct-11B (5\% Forget)}} \\
\midrule
\rowcolor{lightgray}  
Finetune&2.0&2.5&97.7&93.5&90.6&84.3&14.5&16.8&93.7&97.8&62.5&84.3\\

PO~\cite{maini2024tofu}         &89.4&90.0&50.0&{76.9}&27.8&\textbf{84.1}&39.8&36.8&\underline{72.9}&55.5&\underline{43.6}&77.5\\  
GA~\cite{yao2023large}          &89.8&91.1&32.0&29.6&9.7&\underline{80.7}&49.2&47.5&55.1&\underline{79.7}&{35.8}&81.2\\      
GD~\cite{liu2022continual}      &\underline{96.5}&\underline{97.9}&\textbf{82.5}&\underline{77.6}&\textbf{91.0}&\textbf{84.1}&\underline{50.1}&47.5 &\textbf{90.5}&\textbf{91.2}&\textbf{62.2}&\underline{82.4}\\  
KL~\cite{yao2024machine}        &\textbf{100.0}&\textbf{100.0}&23.5&56.3 &39.4&67.8 &47.7&\textbf{53.2}&58.3&17.2&16.8&80.9 \\ 
DPO~\cite{rafailov2024direct}   &83.7&84.8&\underline{52.8}&\textbf{91.5}&\underline{62.5}&80.0&\textbf{52.3}&\underline{52.9}&{62.8}&36.2&32.7&\textbf{83.0}\\ 
\rowcolor{pink}  
Goal
&100.0&100.0&98.1&93.9&91.1&85.2&86.3&87.1&93.9&98.8&63.1&85.2\\
\midrule
\multicolumn{13}{c}{\textbf{LLama-3.2-Vision-Instruct-11B (10\% Forget)}} \\
\midrule
\rowcolor{lightgray}  
Finetune&1.9&2.0&98.1&93.9&90.6&84.3&17.5&16.3&92.9&96.8&63.4&84.3\\

PO~\cite{maini2024tofu}         &84.3&83.7&\textbf{48.7}&\underline{80.8}&\textbf{72.5}&78.8&42.3&42.6&\underline{67.9}&67.2&\underline{46.5}&66.7 \\ 
GA~\cite{yao2023large}          &86.2&85.5&31.6&26.3&8.9&80.3&\underline{57.3}&\underline{57.4}&49.7&\underline{73.7}&35.2&80.5\\  
GD~\cite{liu2022continual}      &\textbf{99.0}&\textbf{100.0}&29.2&51.4&\underline{51.2}&\underline{85.1}&\textbf{59.2}&\textbf{58.6}&\textbf{83.6}&\textbf{79.7}&\textbf{60.8}&\underline{82.1} \\ 
KL~\cite{yao2024machine}        &\underline{96.0}&\underline{96.4}&37.7&49.5&31.8&\textbf{85.4}&51.0&50.7&20.2&50.2 &30.1&75.6 \\ 
DPO~\cite{rafailov2024direct}   &84.1&83.6&\underline{46.4}&\textbf{85.3}&45.4&78.5&50.3&50.6&35.8&22.6&24.6&\textbf{82.8} \\
\rowcolor{pink}  
Goal
&100.0&100.0&98.8&94.2&91.5&85.0&85.6&88.1&94.6&97.6&64.6&85.0 \\
\midrule
\multicolumn{13}{c}{\textbf{LLama-3.2-Vision-Instruct-11B (15\% Forget)}} \\
\midrule
\rowcolor{lightgray}  
Finetune&1.0&1.0&98.9&92.2&90.6&84.3&14.9&16.5&92.6&97.5&63.0&84.3\\

PO~\cite{maini2024tofu}         &85.0&82.0&32.5&67.3&43.5&\underline{80.1}&42.0&41.2&\textbf{38.6}&54.6&\underline{51.7}&58.1 \\  
GA~\cite{yao2023large}          &85.3&84.8&22.6&25.6&8.3&78.3&45.4&45.7&17.6&\textbf{88.5}&19.7&81.5 \\
GD~\cite{liu2022continual}      &\underline{94.3}&\underline{91.1}&\textbf{69.5}&\textbf{89.4}&\textbf{91.0}&79.4&\underline{47.6}&46.8&\underline{33.4}&\underline{86.5}&\textbf{60.4}&\underline{82.7} \\  
KL~\cite{yao2024machine}        &\textbf{96.1}&\textbf{97.9}&16.3&36.2&\underline{61.7}&\textbf{82.9}&45.9&\textbf{47.6}&12.5&18.7&23.7&80.3 \\ 
DPO~\cite{rafailov2024direct}   &87.4&88.2&\underline{37.5}&\underline{80.1} &44.8&77.6&\textbf{48.1}&\underline{47.2}&20.7&14.4&16.5&\textbf{83.2} \\
\rowcolor{pink}  
Goal
&100.0&100.0&99.0&94.6&92.2&86.7&85.3&86.4&93.8&98.4&63.8&86.7\\
\bottomrule
\end{tabular}%
}

\vspace{2mm}
\label{tab:main}
\end{table*}

\subsection{Experiment Results}
\label{sec:Experiment}

We evaluate 5 baseline unlearning methods for each unlearning target (person and event). Recognizing the inherent trade-off between forgetting efficacy and model utility, we adopt an early stopping strategy guided by training loss, following the practices outlined in FIUBench~\cite{ma2024benchmarking}.

\begin{figure*}[!t]
\vspace{10pt}
\centering
  \resizebox{1\linewidth}{!} {
    \includegraphics{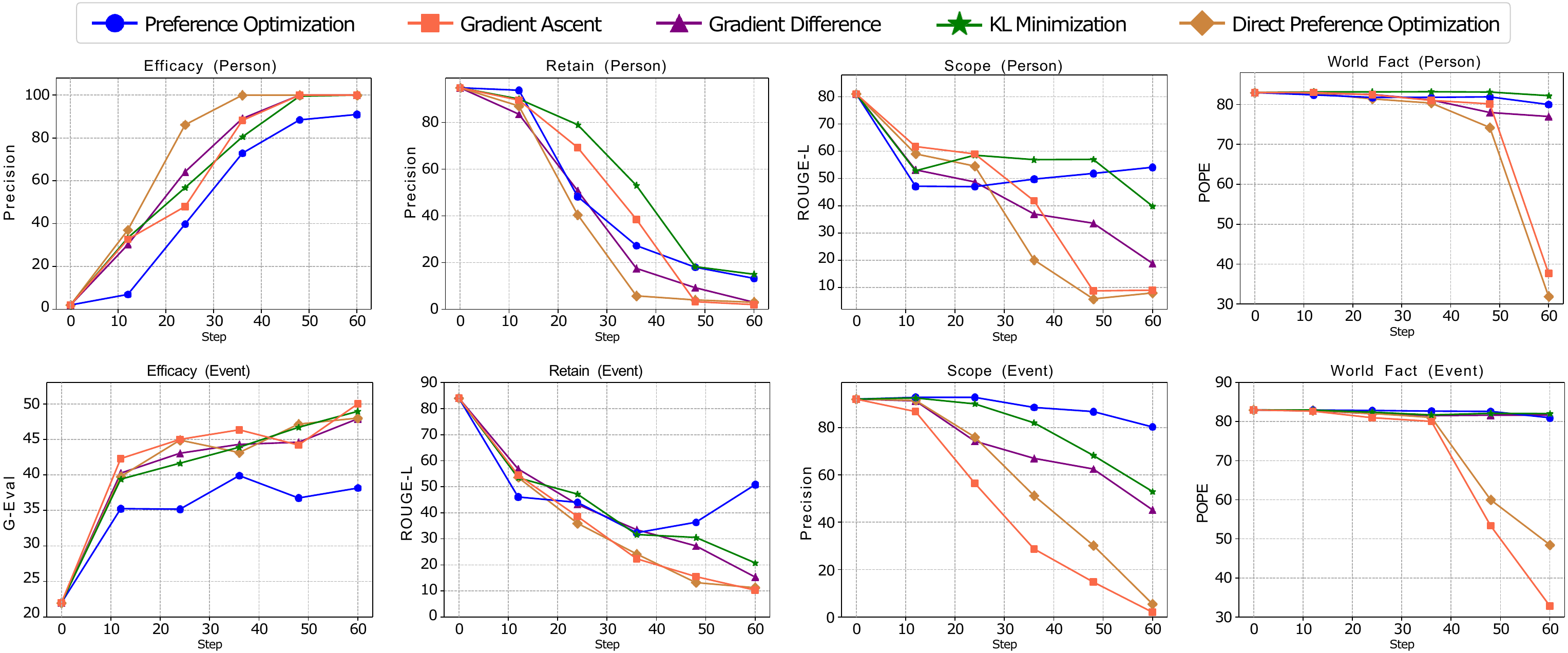}
  }
  \vspace{-10pt}  
  \caption{Performance comparison of various unlearning methods under LLaVA-1.5-7B with a 5\% forget set, evaluated over different unlearning steps.}
    \label{fig:steps}
  \vspace{2mm}
\end{figure*}

\begin{figure*}[!t]
\centering
  \resizebox{1\linewidth}{!} {
    \includegraphics{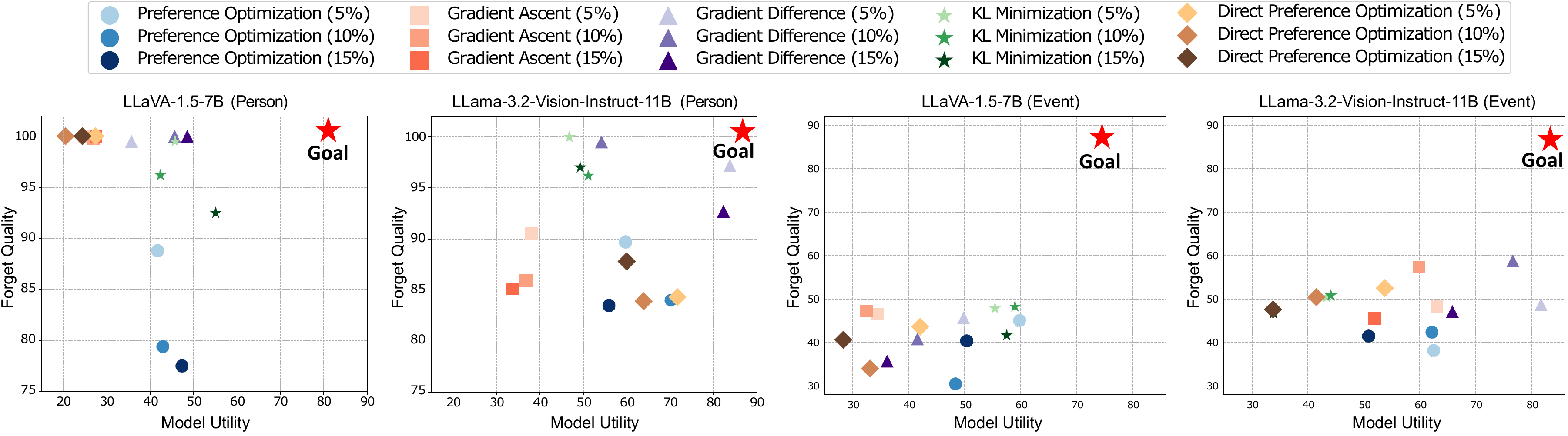}
  }
  \vspace{-10pt} 
  \caption{Overall trade-off between forget quality and model utility across all unlearning baselines using different forget set sizes on LLaVA-1.5-7B and LLaMA-3.2-Vision-11B. The x-axis represents the average of Retain, Scope, Real, and World Fact metrics, indicating model utility. The y-axis represents the average of Efficacy and Generality, reflecting forget quality.}
    \label{fig:trade-off}
  \vspace{5pt}  
\end{figure*}

Table~\ref{tab:main} presents the quantitative results on PEBench across all methods, using the evaluation metrics defined in Section~\ref{sec:metri}. The evolution of forgetting performance over training steps is depicted in Fig.~\ref{fig:steps}, while Fig.~\ref{fig:trade-off} illustrates the trade-off between forgetting quality and model utility. Key observations are summarized below.

\textit{1) Effectiveness of PEBench:} As shown in Table~\ref{tab:main}, while most methods achieve near-perfect efficacy in person unlearning, the performance for event unlearning varies substantially across methods. This disparity \textbf{underscores the importance of incorporating event-level metrics into MU evaluation frameworks} to ensure comprehensive assessment.

Moreover, the scope of unlearning remains a critical but often underexplored dimension~\cite{liu2025rethinking}. When unlearning person entities, the ROUGE-L score for related event descriptions declines sharply from an average of 88.6 to 46.2. When forgetting event scenes, the accuracy of person recognition drops from 96.4\% to an average of 55.2\%. \textbf{These findings demonstrate that unlearning one visual concept can impair the model's ability to recognize related concepts within the same image}, a form of \textbf{cross-concept interference} not captured by prior benchmarks. PEBench addresses this gap by explicitly coupling person and event information, enabling systematic evaluation of such interactions.

\textit{2) Limitations of Current Methods:} For person unlearning, most methods achieve high efficacy and generality; however, the retention of non-target knowledge (Retain and Real sets) often suffers, with the most pronounced degradation observed in the Gradient Ascent (GA) method. In contrast, Gradient Difference (GD) better preserves performance on these sets, suggesting that \textbf{balancing the loss contributions between the Forget and Retain sets is essential}.

For event unlearning, methods based on KL-divergence minimization consistently demonstrate superior efficacy and generality, underscoring their strength in managing semantically rich visual information. This regularization is beneficial in the multimodal setting, where event representations often involve complex interactions between visual and textual.

Among alignment-based approaches, Direct Preference Optimization (DPO) consistently surpasses Preference Optimization (PO) in forgetting efficacy, particularly for person unlearning. This highlights that \textbf{amplifying the reward signal associated with preference violations significantly enhances the model’s ability to unlearn targeted content}.

\textit{3) Impact of unlearning steps:} As shown in Fig.~\ref{fig:steps}, increasing the number of unlearning steps consistently improves forgetting efficacy across all methods, but often at the expense of model utility. This trade-off is especially pronounced in the person unlearning setting, where utility metrics such as Retain and Scope degrade more sharply than in the event unlearning scenario. This suggests that controlling the unlearning trajectory, through step-wise optimization, can significantly influence the balance between forgetting targeted knowledge and preserving overall model performance.

Among all methods, Preference Optimization and KL Minimization maintain superior utility across unlearning steps. Notably, Preference Optimization even yields modest gains in specific utility metrics such as Scope (Person) and Retain (Event), highlighting its capacity to achieve targeted forgetting with minimal utility degradation. These results underscore the promise of preference based approaches in striking a favorable balance between unlearning efficacy and model utility.

\textit{4) Trade-off between forget quality and model utility:} 
As shown in Fig.~\ref{fig:trade-off}, a clear trade-off exists between forget quality and model utility across all methods and model scales. This trade-off is particularly evident when comparing different model architectures. Notably, LLaMA-3.2-Vision-Instruct-11B consistently outperforms LLaVA-1.5-7B in terms of model utility, indicating that larger model capacity aids in better retention of non-targeted knowledge. The larger capacity of LLaMA-3.2-Vision-11B enables it to better preserve general knowledge while forgetting targeted concepts, suggesting that model scale plays a critical role in mitigating the collateral damage of unlearning.

Across all configurations, the goal model (upper-right corner) serves as an ideal reference point, demonstrating the highest achievable forget quality and utility. Among the evaluated baselines, Gradient Difference and KL Minimization demonstrate stronger performance, achieving a more favorable balance between forgetting and utility. Specifically, Gradient Difference applies gradient descent on the retain set to offset forgetting-induced drift, preserving non-targeted knowledge. KL Minimization, by regularizing outputs toward a neutral reference, maintains semantic structure, highlighting the need for precision and stability in unlearning optimization.

\textit{5) Impact of forget set splits:} Following the evaluation of MLLMU~\cite{liu2024protecting}, we divide the benchmark into three different forget set proportions: 5\%, 10\%, and 15\%, while the corresponding retain sets consist of the remaining 95\%, 90\%, and 85\%, respectively. As illustrated in Fig.~\ref{fig:trade-off}, a distinct trend emerges from this setup. As the size of the forget set increases, the overall performance of all evaluated methods drifts progressively further from the ideal performance of the goal model. This drift signifies a more severe trade-off between forgetting quality and model utility at larger scales. These findings underscore the scalability challenges of Machine Unlearning (MU) in Multimodal Large Language Models (MLLMs). 


\begin{figure*}[!t]
\vspace{-15pt} 
\centering
  \resizebox{1\linewidth}{!} {
    \includegraphics{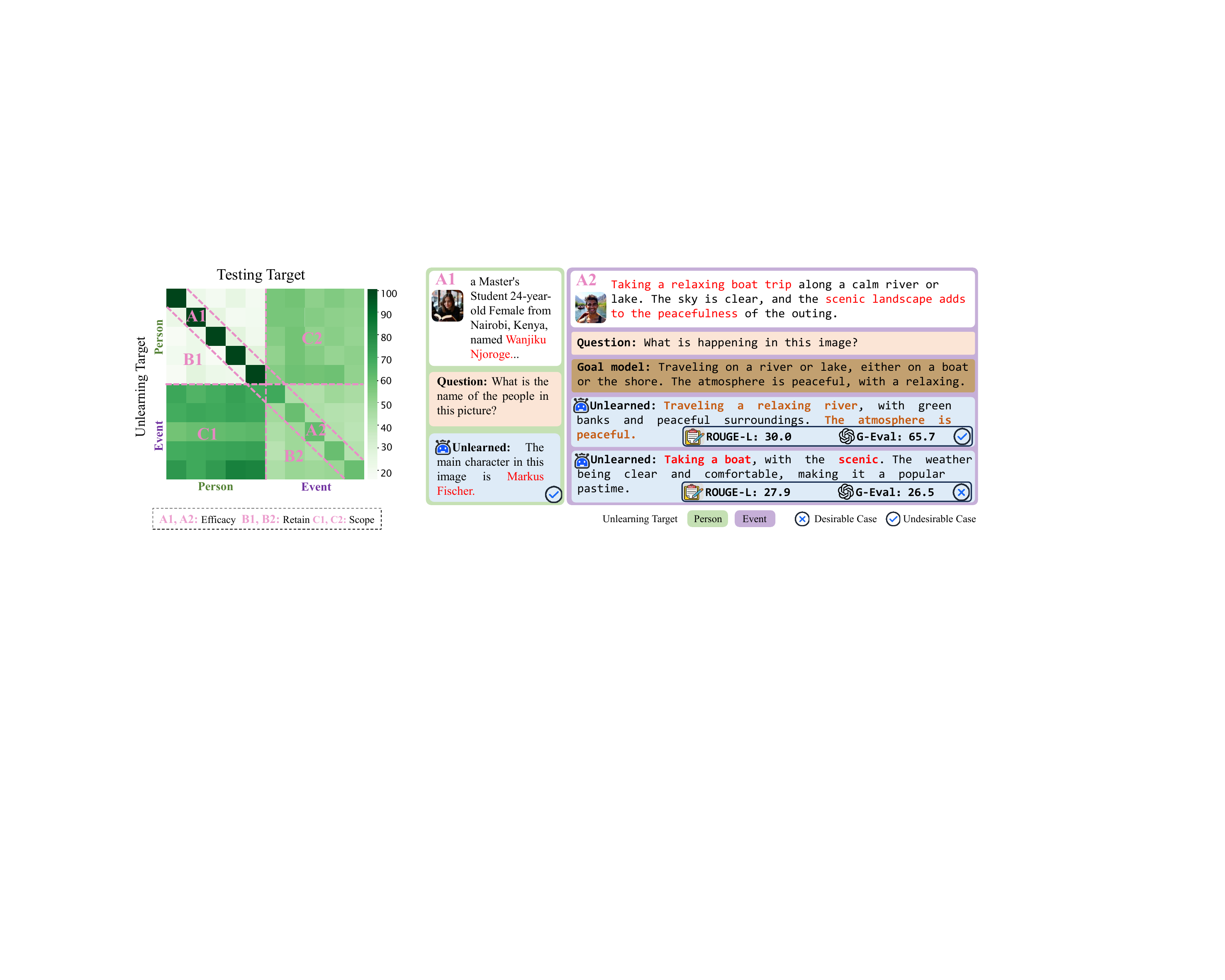}
  }
  \caption{Visualization of GD~\cite{liu2022continual} on LLaVA-1.5-7B with a 5\% forget set.  
\textbf{Left:} Heatmap showing diagonal efficacy and off-diagonal retain/scope effects for person and event targets. The x-axis represents the tested targets, and the y-axis shows the corresponding unlearning targets. Regions A (efficacy), B (retain), and C (scope) are further divided by target type (1 = Person, 2 = Event).  
\textbf{Right:} Examples illustrating unlearning efficacy for both person and event targets.}
    \label{fig:sample}
\vspace{-5pt}  
\end{figure*}

\subsection{Visualization}
Fig.~\ref{fig:sample} provides a comprehensive visualization of unlearning behavior on PEBench. Several key insights can be observed:

\textit{1) Contrasting behaviors for person vs. event unlearning.} The visualization reveals distinct behaviors for person and event unlearning. In the heatmap (Fig.~\ref{fig:sample} left), the deeper diagonal shade for person targets (e.g., A1) compared to event targets (e.g., A2) indicates that models forget individual identities more easily. However, the off-diagonal entries (e.g., B1, B2) show that person unlearning induces greater collateral damage to unrelated knowledge, particularly within the retain set. In contrast, event unlearning yields lower efficacy but better preserves broader model utility. These findings underscore the importance of treating person and event unlearning as distinct challenges in MLLMs.

\textit{2) Effectiveness of G-Eval.}
Successful unlearning requires the model’s output to diverge from the fine-tuned model (which retains the target) and align with the goal model (which has never seen the target). As illustrated in the event unlearning example in Fig.~\ref{fig:sample} (right), G-Eval effectively captures this dual objective by assigning high scores when both criteria are satisfied. In contrast, traditional lexical metrics like ROUGE-L are less sensitive to such semantic shifts, highlighting the effectiveness of LLM-based evaluation.

\subsection{Simultaneously Unlearning Both Targets}

\textit{Experiment setting.} We extend the evaluation to the scenario of simultaneously unlearning both person and event, using a 5\% forget set for each target. Based on their superior performance in Table~\ref{tab:main}, we select GD and KL as representative methods. During training, models are guided to generate incorrect responses regarding both names and events. During testing, the efficacy of unlearning is assessed separately for each forget target.

As shown in Table~\ref{tab:two_target}, jointly unlearning both concepts leads to a noticeable decline in performance across both targets. For instance, the person unlearning efficacy drops significantly (from an average of 99.0 to 86.2), while event unlearning efficacy declines from 47.8 to 45.1. These results highlight the intrinsic difficulty of forgetting joint concepts.

\begin{figure}[tb] \centering
    \includegraphics[width=0.42 \textwidth]{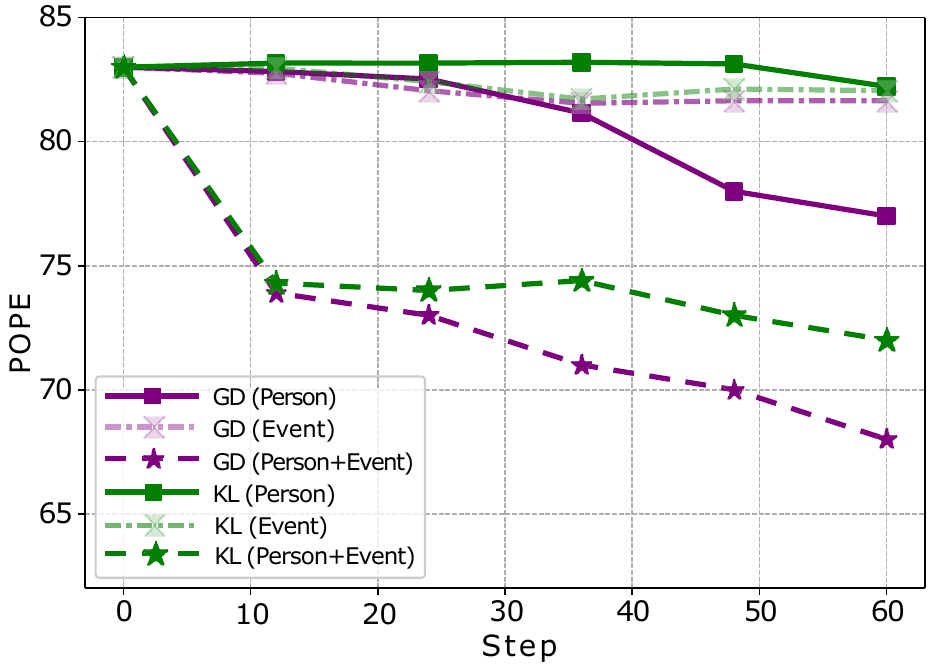}
\vspace{-10pt}    
\caption{Comparison of model utility (world fact) degradation under different unlearning targets using LLaVA-1.5-7B with a 5\% forget set.} 

\label{fig:both_t} 
\vspace{-15pt}
\end{figure}

\begin{table*}[t]
\vspace{-15pt} 
\centering
\caption{Performance overview of simultaneously unlearn person and event. $\textcolor{Highlight}{+}$ (or $\textcolor{red}{-}$) indicates the performance gain (or decrease) compared to the base method.
}
\vspace{-5pt} 
\resizebox{\textwidth}{!}{%
\begin{tabular}{c|lllll|lllll}

\midrule
\multirow{2}{*}{\textbf{Method}}&\multicolumn{5}{c|}{\textbf{Person Unlearning}} & \multicolumn{5}{c}{\textbf{Event Unlearning}} \\
\cline{2-11}
& \textbf{Efficacy} & \textbf{Generality} & \textbf{Retain} & \textbf{Real}  & \textbf{World Fact}  &\textbf{Efficacy}  & \textbf{Generality}  & \textbf{Retain} & \textbf{Real} & \textbf{World Fact} \\
\midrule
\multicolumn{11}{c}{\textbf{LLaVA-1.5-7B}} \\
\midrule
\rowcolor{lightgray}Finetune&2.0&1.5&94.8 &67.7&82.3&18.0&21.3&84.4&39.6&82.3 \\
GD~\cite{liu2022continual}&{86.9}&{86.0}&{13.0}&{28.6}&{71.4} &{44.3}& 39.2&21.0&16.8&{71.4}  \\

GD$_{+BGD}$&93.2$_{\hl{+6.3}}$&92.5$_{\hl{+6.5}}$&11.2$_{\textcolor{red}{-1.8}}$&54.3$_{\hl{+25.7}}$&77.3$_{\hl{+5.9}}$&49.2$_{\hl{+4.9}}$&46.6$_{\hl{+7.4}}$&23.4$_{\hl{+2.4}}$&24.7$_{\hl{+7.9}}$&76.3$_{\hl{+4.9}}$ \\
KL~\cite{yao2024machine}&{72.9}&{71.1}&{7.8}&{37.9}&{75.5} &{43.9}& 40.4&23.2&29.6&{75.5} \\

KL$_{+BGD}$&89.5$_{\hl{+16.6}}$&88.2$_{\hl{+17.1}}$&18.6$_{\hl{+10.8}}$&48.3$_{\hl{+10.4}}$&80.1$_{\hl{+4.6}}$&49.6$_{\hl{+5.7}}$&50.0$_{\hl{+9.6}}$&28.4$_{\hl{+5.2}}$&27.3$_{\textcolor{red}{-2.3}}$&80.1$_{\hl{+4.6}}$ \\

\rowcolor{pink}Goal
&100.0&100.0&95.0& 71.8  &82.8&91.9&86.8&85.6&40.3&82.8 \\

\midrule
\multicolumn{11}{c}{\textbf{LLama-3.2-Vision-Instruct-11B}} \\
\midrule
\rowcolor{lightgray}Finetune&2.0&2.5&97.7 &90.6&84.3&14.5&16.8&93.7&62.5&84.3 \\
GD~\cite{liu2022continual}&{95.0}&{91.0}&{37.5}&{89.5}&{82.8} &{45.8}& 47.2&80.2&25.0&{82.8 }  \\
GD$_{+BGD}$&96.2$_{\hl{+1.2}}$&94.5$_{\hl{+3.5}}$&52.8$_{\hl{+15.3}}$&76.5$_{\textcolor{red}{-13.0}}$&83.6$_{\hl{+0.8}}$&46.8$_{\hl{+1.0}}$&47.7$_{\hl{+0.5}}$&72.5$_{\textcolor{red}{-7.7}}$&54.3$_{\hl{+29.3}}$&81.8$_{\textcolor{red}{-1.0}}$ \\

KL~\cite{yao2024machine}&{90.0}&{89.5}&{18.5}&{67.3}&{56.8} &{46.4}& 46.1&21.0&24.3&{56.8} \\
KL$_{+BGD}$&92.4$_{\hl{+2.4}}$&92.2$_{\hl{+2.7}}$&35.2$_{\hl{+16.7}}$&50.6$_{\textcolor{red}{-16.7}}$&77.4$_{\hl{+20.6}}$&45.9$_{\textcolor{red}{-0.5}}$&47.5$_{\hl{+1.4}}$&66.3$_{\hl{+45.3}}$&60.1$_{\hl{+35.8}}$&72.9$_{\hl{+16.1}}$ \\

\rowcolor{pink}Goal
&100.0&100.0&98.1& 91.1  &85.2&86.3&87.1&93.9&63.1&85.2 \\
\bottomrule[1pt]
\end{tabular}%
}
\vspace{-2pt} 

\label{tab:two_target}
\end{table*}
There are two key challenges in jointly unlearning. First, two unlearning targets present inherently conflicting optimization objectives within a single image: unlearning a person requires removing identity-specific features, whereas unlearning an event demands erasing scene-level semantics. This conflict hinders the model's general utility, as reflected by the drop in the World Fact metric (see Fig.~\ref{fig:both_t}). Second, an inherent data imbalance exists, where each event is associated with multiple individuals. This arrangement creates a strong entanglement that degrades unlearning performance, particularly for person entities, as indicated by the Scope metric in Table~\ref{tab:main}.

To address this challenge, we propose \textbf{Balanced Gradient Difference (BGD)}, a method incorporating both data-level and task-level balancing. At the \textit{data level}, we mitigate the inherent imbalance between person and event samples by dynamically adjusting the sampling ratio. Specifically, event-related forget samples are initially excluded and then progressively introduced by increasing their proportion by 5\% at each training step. Additionally, we apply randomized sampling within each class to maintain intra-class diversity. At the \textit{task level}, we decompose the total loss into distinct components corresponding to each unlearning target:

\vspace{-5pt}
\begin{equation}
\scriptsize
    L_{\text{BGD}} = - \alpha \cdot L(\mathcal{D}_{\text{person}}^f, w) - \beta \cdot L(\mathcal{D}_{\text{event}}^f, w) + \gamma \cdot  L(\mathcal{D}^r, w),
\end{equation}
where \(\alpha\), \(\beta\), and \(\gamma\) are weights, which we set by default to 0.3, 0.2, and 0.5 based on task difficulty.

Empirical results in Table~\ref{tab:two_target} demonstrate that BGD significantly improves performance, particularly for person unlearning, validating its effectiveness in mitigating the conflicts arising from joint concept forgetting. This setting mirrors real-world use cases such as removing fake news, where both person and event information need to be erased concurrently.

\section{Discussion}

\subsection{Findings and Challenges in Multimodal Unlearning}
Our study uncovers several key findings and challenges in multimodal unlearning. First, person unlearning tends to achieve higher forgetting efficacy but incurs greater degradation in model utility, whereas event unlearning shows the opposite trend. Second, forgetting one visual concept (e.g., a person) can unintentionally impair the model’s performance on semantically coupled concepts (e.g., events). These two findings highlight the need for more disentangled representations to enable precise and targeted unlearning. Third, as the size of the forget set increases, model performance deteriorates, revealing scalability limitations in existing MU methods. Finally, simultaneously unlearning both persons and events introduces conflicting optimization objectives, which we mitigate through the proposed Balanced Gradient Difference (BGD) method. This suggests that adaptive strategies, such as curriculum-based, may be necessary to balance forgetting efficacy with utility preservation in MLLMs.

\subsection{Limitations and Future Work}
Despite the comprehensive design of PEBench, several limitations remain. First, the benchmark focuses on forgetting visual concepts within static images, whereas MLLMs operate across richer modalities such as video and audio, which are not yet included. Extending MU evaluation to these modalities is an important direction for future research. Second, although our proposed BGD method alleviates conflicts when forgetting persons and events, it still underperforms on world fact metrics. This underscores the challenge of simultaneously unlearning coupled visual concepts. In future work, we aim to expand PEBench to encompass broader modalities and investigate more robust MU strategies tailored to the unique characteristics of MLLMs.

\section{Conclusion}
We presented PEBench, a comprehensive benchmark for evaluating machine unlearning (MU) in multimodal large language models (MLLMs), with a focus on both personal identities and event scenes. Our results underscore the necessity of diverse unlearning targets, revealing that while most methods excel in forgetting individuals, event-related unlearning remains more challenging and variable. By enabling systematic, multi-dimensional evaluation, PEBench provides a rigorous framework for advancing MU research in MLLMs and highlights the importance of target-specific strategies for effective and reliable unlearning.

\bibliographystyle{IEEEtran}
\bibliography{main}

\end{document}